\documentclass{article}
\usepackage{fancyhdr}

\fancypagestyle{arxivfirst}{
    \fancyhf{}
    \fancyhead[L]{%
        \parbox[b]{\textwidth}{\small\bfseries
        Cite as: Sijie Xu and Francisco Cruz.
        Integrating Action Prefix Prediction and Alignment Across Modalities
        for Multimodal Reinforcement Learning.
        \textit{Proceedings of the Australasian Conference on Robotics and
        Automation (ACRA), 2026.}}%
    }
    
}
\usepackage{acra}
\usepackage{graphicx}
\usepackage{amsmath}
\usepackage{amssymb}
\usepackage{balance}
\title{Integrating Action Prefix Prediction and Alignment Across Modalities
for Multimodal Reinforcement Learning}
\author{
Sijie Xu$^{1}$ \qquad Francisco Cruz$^{1,2}$ \\
$^{1}$
School of Computer Science and Engineering, UNSW Sydney, Australia \\
$^{2}$ Escuela de Ingeniería,
Universidad Central de Chile, Santiago, Chile \\
\texttt{sijie.xu@student.unsw.edu.au, f.cruz@unsw.edu.au}
}

\begin{document}

\maketitle
\thispagestyle{arxivfirst}

\begin{abstract}

Learning representations from multimodal observations is difficult in
reinforcement learning with limited interaction data. Multi-Modal Joint
Predictive Representation (MJPR) combines multimodal latent prediction with
contrastive learning across modalities, but its recursive rollout depends on
earlier predictions and its pairwise alignment focuses on correspondence
between modalities. We propose Improved Predictive Representation (IPR),
which combines prediction from action prefixes with a Contrastive Multimodal
learning (CoMM) objective applied at the observed anchor. At each horizon,
IPR predicts future representations for each modality directly from a common
observed anchor and the corresponding action sequence. We evaluate
\textsc{IPR-S} with a minimum retained crop area of $10\%$ and
\textsc{IPR-H} with $50\%$. Across five DeepMind Control Suite tasks,
\textsc{IPR-H} achieves an average return $10.0\%$ higher
than the MJPR baseline
with visual and depth images and $3.4\%$ higher with visual images and
position vectors. The relative performance of the two crop settings varies
across tasks. In Webots navigation, both IPR configurations achieve higher returns than the baseline without LiDAR, while the baseline achieves the highest return when LiDAR is available.

\end{abstract}
% ------------------------------------------------------------------
% 1. Introduction
% ------------------------------------------------------------------

\section{Introduction}
\label{sec:introduction}
Reinforcement learning (RL) agents learn control policies through interaction
with an environment \cite{sutton2018reinforcement}. Multimodal control may
combine images with depth, range, or low-dimensional state observations.
These sources provide complementary information, but limited interaction
data and reward supervision make it difficult to learn how to use them
together.

Predicting representations of future observations from actions provides an
additional learning signal. Self-Predictive Representations (SPR) uses this
idea \cite{schwarzer2021spr}, and Multi-Modal Joint Predictive Representation (MJPR) extends it with modality-specific latent prediction and cross-modal contrastive learning \cite{wang2025mjpr}. MJPR predicts future
representations recursively, so errors can affect later horizons and the predictions must be generated in sequence. Its pairwise alignment also emphasises information shared between modalities.

Predicting each horizon from action prefix offers an alternative to MJPR's recursive rollout. Fast LeWorldModel uses this approach for offline planning with a single visual modality \cite{gao2026fastlewm}. For multimodal alignment, Contrastive Multimodal learning (CoMM) relates fused representations to views retaining individual modalities \cite{dufumier2025comm}. Its input transformations, however, have no direct
equivalent for predicted latent representations.

\begin{figure*}[t]
    \centering
    \includegraphics[width=0.96\textwidth]
    {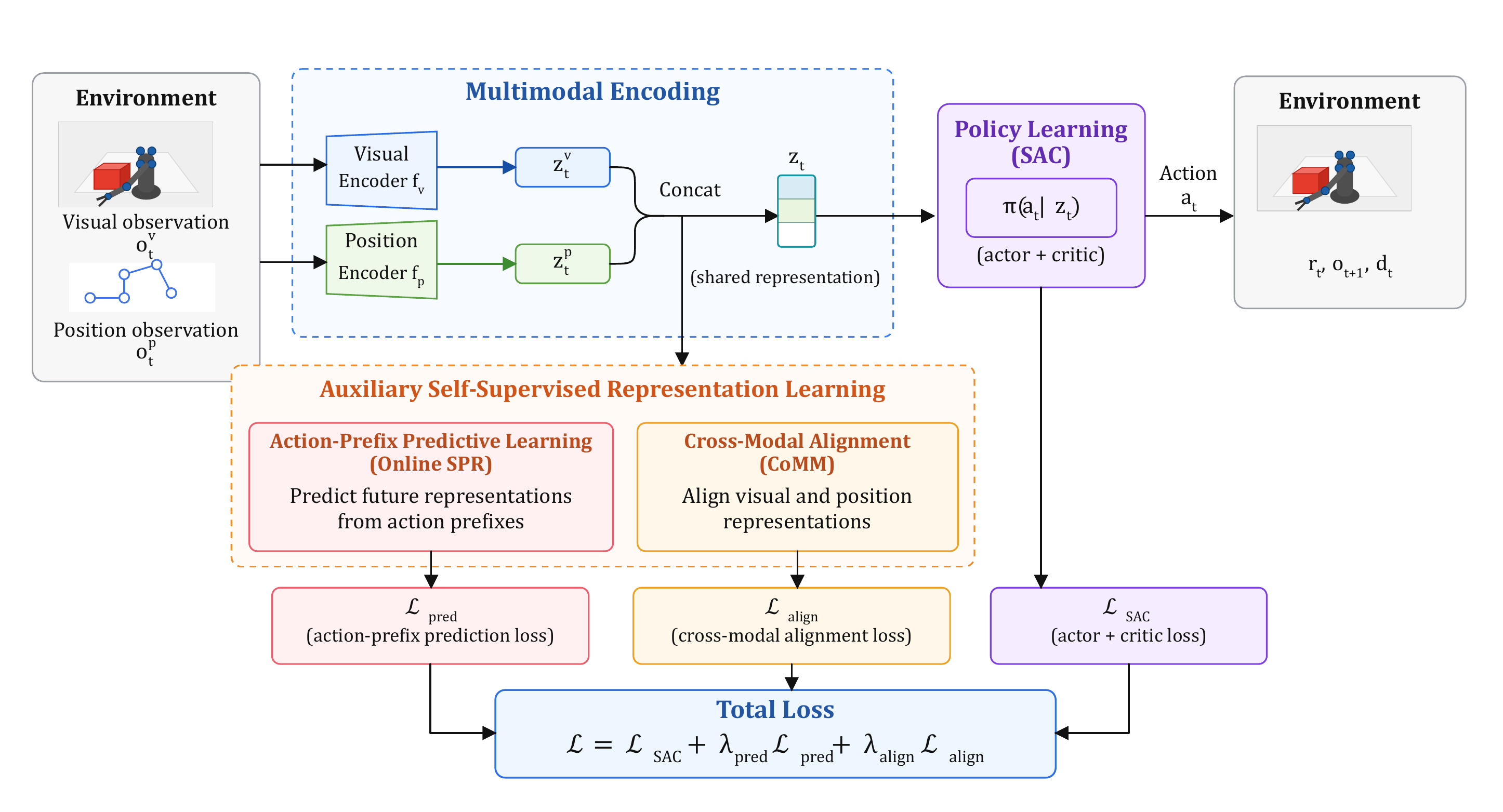}
    \caption{High-level overview of IPR using visual and position observations
    as an example. The encoders provide representations to SAC and the two
    auxiliary branches. Action-prefix prediction provides temporal
    supervision, while CoMM style alignment relates complete and
    modality-retaining views at the anchor observation. Superscripts $v$ and
    $p$ denote visual and position representations.}
    \label{fig:ipr-high-level}
\end{figure*}

We integrate these ideas into MJPR as \emph{Improved Predictive
Representation} (IPR) for online multimodal control. IPR predicts each modality's future representation directly from the same observed anchor and the corresponding action prefix. It applies CoMM style alignment to augmented anchor observations rather than predicted representations. The objectives train modality encoders shared with a Soft Actor--Critic (SAC) agent \cite{haarnoja2018sac}. Figure~\ref{fig:ipr-high-level} shows the
framework.

We evaluate IPR on DeepMind Control Suite (DMC) tasks
\cite{tassa2018deepmind} using RGB images paired with depth images or
position vectors, and in Webots navigation \cite{michel2004webots}.
Both IPR configurations have higher observed averages than the MJPR
baseline in the DMC settings. In Webots, the baseline performs better
with LiDAR, whereas IPR has higher returns without it. Component ablations examine the two changes separately. A wall time comparison measures how replacing recursive prediction with action-prefix prediction affects total training time.
% ------------------------------------------------------------------
% 2. Background and Related Work
% ------------------------------------------------------------------

\section{Background and Related Work}
\label{sec:related}

\paragraph{Self-Supervised Representation Learning.}

Soft Actor--Critic (SAC) is an off-policy actor--critic algorithm that
maximises expected return and policy entropy \cite{haarnoja2018sac}. Its
maximum-entropy objective is
\begin{equation}
J(\pi)
=
\mathbb{E}_{\pi}\!\left[
\sum_{t=0}^{\infty}
\gamma^t
\left(
r_t+\alpha\mathcal{H}\bigl(\pi(\cdot\mid s_t)\bigr)
\right)
\right],
\label{eq:sac-objective}
\end{equation}
where $\pi$ is the policy, $s_t$ and $r_t$ are the state and reward at time
$t$, $\gamma$ is the discount factor, $\alpha$ controls the entropy
contribution, and $\mathcal{H}$ denotes policy entropy. SAC provides the
control backbone used by MJPR and IPR.

In image-based RL, Reinforcement Learning with Augmented Data (RAD) and
Data-Regularized Q (DrQ) apply image transformations to regularise
representation and value learning
\cite{laskin2020rad,yarats2021drq}. Contrastive Unsupervised Representations
for Reinforcement Learning (CURL) instead contrasts augmented views of the
same observation \cite{laskin2020curl}. These methods encourage consistency
between augmented observations but do not explicitly model
action-conditioned temporal transitions.

Temporal objectives also use sequential experience. Contrastive Predictive
Coding distinguishes future representations from negative samples
\cite{oord2018cpc}, while Augmented Temporal Contrast (ATC) contrasts
observations sampled from nearby points in an RL trajectory
\cite{stooke2021atc}. Self-Predictive Representations (SPR) predicts future
representations under the intervening actions and matches them to targets
produced by a slowly updated encoder \cite{schwarzer2021spr}.

\paragraph{Multimodal Predictive and Contrastive Learning.}

Multimodal RL combines observations with different structures, including
visual images, depth images, range data, and low-dimensional state vectors.
Modality-specific encoders preserve information suited to each sensor, while
a fused representation allows the policy and value functions to use
complementary information.

MJPR extends SPR to multimodal RL by predicting modality-specific future
representations and applying cross-modal contrastive learning during the
latent rollout \cite{wang2025mjpr}. Its reported average across tasks is
higher than the multimodal SAC, CURL, DrQ, and SPR variants evaluated in the
same study. MJPR is also the closest architectural predecessor to IPR, as
both use SAC, modality-specific encoders, momentum target networks, and
multimodal latent prediction.

MJPR nevertheless uses a recursive latent rollout in which each future prediction
depends on the preceding predicted representation. Its pairwise cross-modal
alignment is also applied throughout the rollout and mainly emphasises
correspondence shared between modalities.

Contrastive learning commonly treats augmented views of the same sample as
positive pairs \cite{chen2020simclr}. Contrastive Multiview Coding extends
this principle across sensory views \cite{tian2020cmc}, while CoMM relates
modality-retaining views to augmented fused representations
\cite{dufumier2025comm}. CoMM is designed to represent redundant, unique, and
synergistic multimodal information. However, its input transformations do
not have an equivalent operation on predicted latent representations. This
creates a difficulty when integrating CoMM with a predictive latent rollout.

\paragraph{Action-Prefix Latent Prediction.}

Fast LeWorldModel replaces repeated one-step rollout with action-prefix
prediction, allowing multiple future representations to be predicted from a
common observation \cite{gao2026fastlewm}. This avoids feeding earlier
predictions back into the transition model and allows the prediction
horizons to be computed in parallel.

Fast LeWorldModel studies offline planning with a single visual modality.
IPR instead applies action-prefix prediction as an auxiliary objective for
online multimodal continuous control and combines it with anchor-level
cross-modal alignment.

% ------------------------------------------------------------------
% 3. Method
% ------------------------------------------------------------------

\section{Method}
\label{sec:method}

IPR modifies the multimodal predictive framework of MJPR
\cite{wang2025mjpr} in two ways. It replaces recursive latent rollout with
prediction conditioned on action prefixes and replaces pairwise alignment
during the rollout with anchor-level cross-modal alignment.

\subsection{Problem Formulation}
\label{sec:problem-formulation}

At time $t$, the agent receives a multimodal observation
$o_t=\{o_t^m\}_{m\in\mathcal{M}}$, where $\mathcal{M}$ is the set of
modalities and $m$ indexes a modality. After selecting action $a_t$, the
agent receives reward $r_t$ and the next observation $o_{t+1}$. Each
modality-specific encoder $f_m$ maps its observation to the latent
representation $z_t^m=f_m(o_t^m)$. The modality representations are
concatenated and supplied to the SAC actor and critics. The auxiliary
objectives sample contiguous sequences from the replay buffer, using $o_t$
as the anchor observation and later observations as prediction targets.
\subsection{IPR Framework}
\label{sec:framework}

\begin{figure*}[t]
    \centering
    \includegraphics[width=0.95\textwidth]{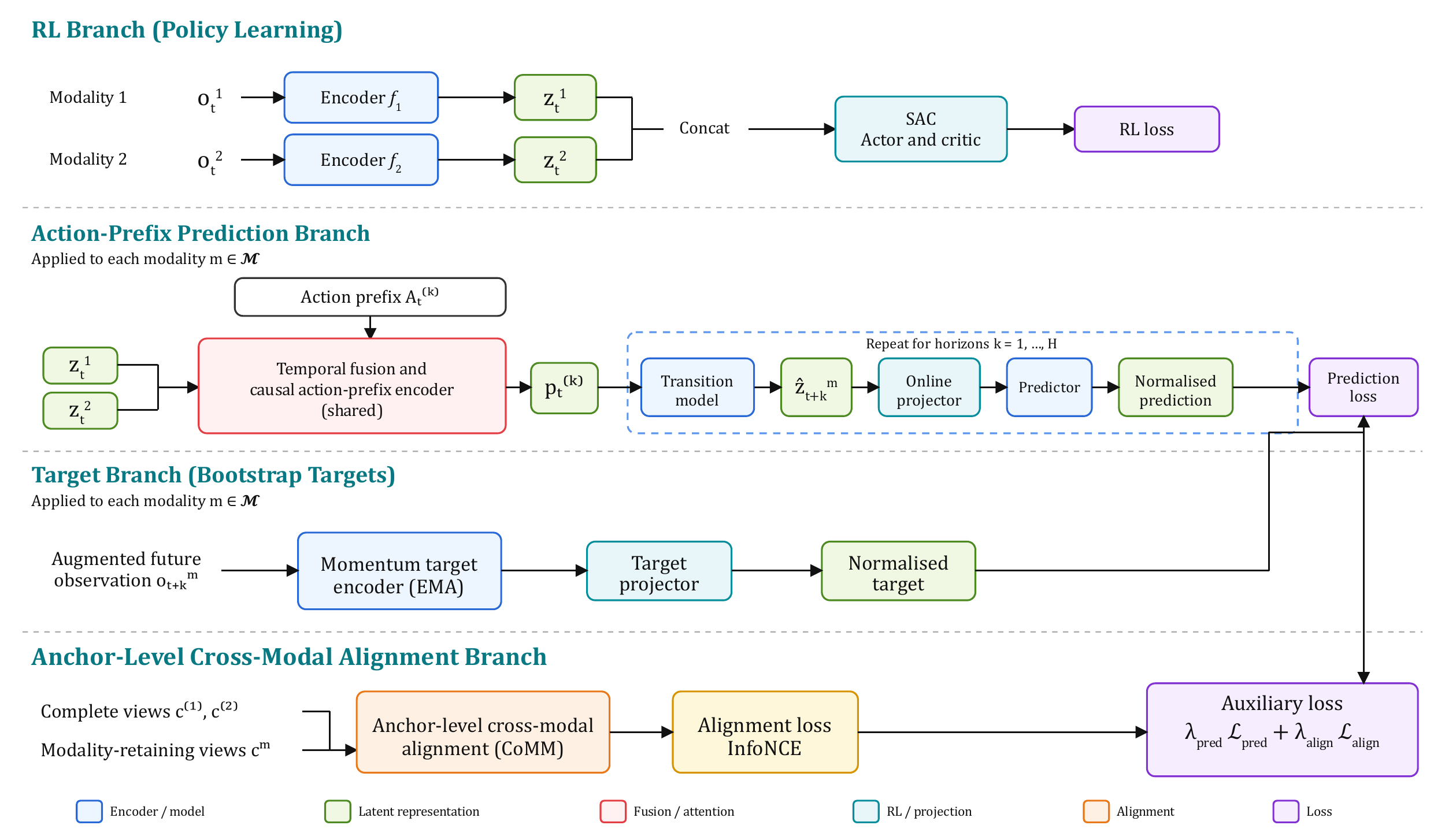}
    \caption{Overview of IPR. The prediction branch predicts
    modality-specific future representations from a common anchor, while
    momentum target encoders provide stop-gradient targets. The alignment
    branch contrasts complete multimodal views with modality-retaining views
    at the anchor observation. An example with two modalities is shown.}
    \label{fig:ipr-architecture}
\end{figure*}

Figure~\ref{fig:ipr-architecture} illustrates the two auxiliary branches.
The action-prefix prediction branch contains a temporal fusion network, an
action-prefix encoder, modality-specific transition models, and momentum
target encoders. The cross-modal alignment branch uses a separate fusion
network, learned modality mask tokens, and a shared projection head. Both
branches update the online modality encoders shared with the SAC agent.

\paragraph{Data Augmentation.}
All transformations are applied to observations before modality encoding.
None is applied directly to a predicted latent representation. In the
action-prefix prediction branch, anchor and target image observations are
augmented using random spatial shifts. For DMC RGB--Position, $10\%$ of the
position vector entries are independently masked. In Webots, spatial shifts
are applied only to the camera input, while the LiDAR and position inputs
remain unchanged in this branch.

The cross-modal alignment branch uses random resized cropping at the anchor
observation. For RGB--Depth, the same crop parameters are applied to the
spatially aligned visual and depth inputs. For RGB--Position, the visual
observation is cropped and the position vector is independently masked. In
Webots, cropping is applied to the camera observation, while $10\%$ of the
LiDAR and position entries are independently masked.
\paragraph{Action-Prefix Prediction.}
Let $H$ denote the prediction horizon. For each
$k\in\{1,\ldots,H\}$, the corresponding action prefix is
\begin{equation}
A_t^{(k)}
=
(a_t,a_{t+1},\ldots,a_{t+k-1}),
\label{eq:action-prefix}
\end{equation}
which contains the actions connecting the anchor observation $o_t$ to the
target observation $o_{t+k}$.

A temporal fusion network first contextualises the modality
representations. The resulting features are mapped to a state token, while
the actions are mapped to action embeddings. A causal Transformer
\cite{vaswani2017attention} encodes the state token and action sequence to
produce one action-prefix representation for each prediction horizon. Its
causal mask prevents a representation from accessing actions beyond its
corresponding prefix.

For each modality, a modality-specific transition model combines the
original anchor representation, the fused anchor features, and the
corresponding action-prefix representation to predict
$\hat z_{t+k}^m$. All future representations are predicted directly from
the same anchor rather than recursively from earlier predictions. The
horizons can therefore be computed in parallel, following the action-prefix
principle of Fast LeWorldModel \cite{gao2026fastlewm}.

\paragraph{Predictive Targets.}

Following SPR \cite{schwarzer2021spr}, each predicted future representation
is processed by online projection and prediction heads. A momentum target
encoder and projection head produce a stop-gradient target from the
corresponding augmented future observation. The prediction and target are
normalised to unit length and compared using cosine distance. These distances
are averaged across modalities and prediction horizons to form the prediction
loss $\mathcal{L}_{\mathrm{pred}}$. The target networks are updated through
exponential moving averages of their corresponding online networks.

\paragraph{Anchor-Level Cross-Modal Alignment.}

The second auxiliary branch uses a CoMM style alignment objective
\cite{dufumier2025comm}.
The images and position vectors are augmented prior to encoding. The predicted representations no longer have the structure of these inputs, so applying the same augmentations to them would have a different meaning. In preliminary
runs, augmenting these representations gave poorer results. We therefore apply alignment at the observed anchor.

Two complete views are generated independently, each view containing all modalities. For each modality, a modality-retaining view preserves that modality and replaces the remaining modalities with learned mask tokens.
A separate fusion network and a shared projection head transform these views into normalised representations.

Let $c^{(1)}$ and $c^{(2)}$ denote the batches of fused representations
from the two complete views, and let $c^m$ denote the batch obtained from
the view retaining modality $m$. For two representation batches $x$ and
$y$, each containing $B$ samples, the directional Information
Noise-Contrastive Estimation (InfoNCE) loss \cite{oord2018cpc} is
\begin{equation}
\ell_{\mathrm{con}}(x,y)
=
-\frac{1}{B}
\sum_{i=1}^{B}
\log
\frac{
\exp\!\left(\operatorname{sim}(x_i,y_i)/T\right)
}{
\sum_{j=1}^{B}
\exp\!\left(\operatorname{sim}(x_i,y_j)/T\right)
},
\label{eq:method-infonce}
\end{equation}
where $B$ is the batch size, $T$ is the temperature, and
$\operatorname{sim}(\cdot,\cdot)$ denotes cosine similarity. The
cross-modal alignment loss is
\begin{equation}
\begin{aligned}
\mathcal{L}_{\mathrm{align}}
={}&
\frac{1}{1+|\mathcal{M}|}
\Bigg[
\ell_{\mathrm{con}}(c^{(1)},c^{(2)})
\\
&\quad+
\frac{1}{2}
\sum_{m\in\mathcal{M}}
\ell_{\mathrm{con}}(c^m,c^{(1)})
\\
&\quad+
\frac{1}{2}
\sum_{m\in\mathcal{M}}
\ell_{\mathrm{con}}(c^m,c^{(2)})
\Bigg].
\end{aligned}
\label{eq:alignment-objective}
\end{equation}
The first term aligns the two complete views, while the remaining terms align
each modality-retaining view with both complete views. The prediction loss
matches each future prediction to one momentum target, whereas the alignment
loss uses other samples in the batch as contrastive negatives.

The complete auxiliary loss is
$\mathcal{L}_{\mathrm{aux}}
=
\lambda_{\mathrm{pred}}\mathcal{L}_{\mathrm{pred}}
+
\lambda_{\mathrm{align}}\mathcal{L}_{\mathrm{align}}$,
where $\lambda_{\mathrm{pred}}$ and $\lambda_{\mathrm{align}}$ control the
contributions of the prediction and alignment losses, respectively. SAC and
auxiliary updates use the same replay data but are performed through separate
gradient steps. During inference, the auxiliary fusion networks, projection
and prediction heads, action-prefix encoder, transition models, and momentum
target networks are discarded. Only the online modality encoders and actor
are retained.
% ------------------------------------------------------------------
% 4. Experiments
% ------------------------------------------------------------------

\section{Experiments}
\label{sec:experiment}

The DMC benchmark \cite{tassa2018deepmind} is evaluated under
RGB--Depth and RGB--Position observation settings. Component ablations
examine the two modifications and their training time, while Webots provides
a secondary navigation experiment. Return results are aggregated over five
seeds. Learning curves show the mean and one standard deviation. The final
return for each seed is the mean over its last five evaluations.

\subsection{DeepMind Control Suite Environments}
\label{sec:dmc}
\begin{figure}[t]
    \centering
    \includegraphics[width=\columnwidth]{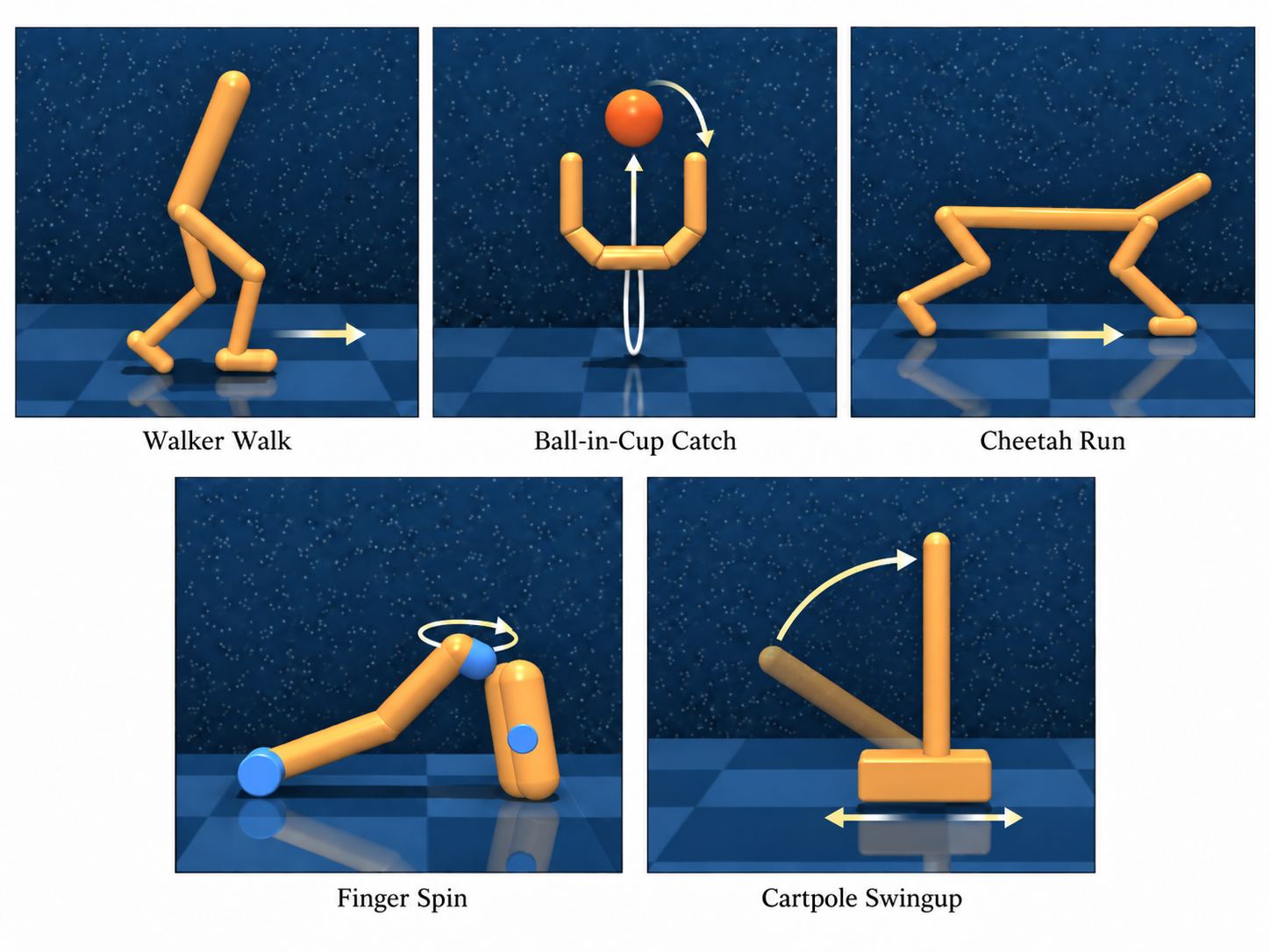}
    \caption{Example RGB camera observations from the five DeepMind Control
    Suite tasks. The added arrows show the main motion directions and are not
    part of the agent observations.}
    \label{fig:dmc-task-examples}
\end{figure}
\paragraph{Experimental Setting.}

We evaluate five continuous control tasks from DMC: Ball-in-Cup Catch,
Cheetah Run, Finger Spin, Cartpole Swing-up, and
Walker Walk. We retain the original task dynamics, reward functions, action
spaces, and termination conditions. DMC provides a reward in $[0,1]$ at each
simulator step, and performance is measured as the sum of rewards over an episode. The task identifiers and action dimensions are listed in
Table~\ref{tab:dmc-tasks}, while Figure~\ref{fig:dmc-task-examples}
illustrates the controlled systems and their main task motions. Actions are
scaled to $[-1,1]$ and repeated for four environment steps.
\begin{table}[t]
    \centering
    \caption{The DMC tasks used in multimodal observation settings, with
    their environment identifiers and action dimensions.}
    \small
    \setlength{\tabcolsep}{3pt}
    \begin{tabular}{@{}lcc@{}}
        \hline
        Task & DMC identifier & Action size \\
        \hline
        Catch    & \texttt{ball\_in\_cup/catch} & 2 \\
        Run      & \texttt{cheetah/run}         & 6 \\
        Spin     & \texttt{finger/spin}         & 2 \\
        Swing-up & \texttt{cartpole/swingup}    & 1 \\
        Walk     & \texttt{walker/walk}         & 6 \\
        \hline
    \end{tabular}
    \label{tab:dmc-tasks}
\end{table}

Both observation settings use three stacked $84\times84$ frames rendered
from camera 0. Although we retain the term RGB when naming the experiments,
the implementation averages the three color channels and normalizes the
resulting single-channel image to $[0,1]$.

We use two observation settings to examine different forms of multimodal
input. RGB--Depth follows the setting used in the original MJPR experiments
\cite{wang2025mjpr}. The visual and depth images are rendered from the same
camera at the same spatial resolution. Complementary diagonal masks remove
$50\%$ of the spatial information from each modality. Each image therefore
contains a different half of the observation, and neither modality is
complete on its own.

RGB--Position combines the visual observation with a position vector
containing the simulator's generalized position coordinates provided by DMC
\cite{tassa2018deepmind}. This setting examines learning from an image and a
lower-dimensional position vector with a different structure and scale. It does
not apply the fixed visual masks used in RGB--Depth.

We denote our MJPR implementation by \textsc{Base}
\cite{wang2025mjpr}. Its optimisation corrects the repeated update of shared online encoders by excluding them from the prediction module's optimiser. Their auxiliary gradients are applied by the dedicated encoder optimisers.

The baseline and IPR use the same optimization
implementation, SAC backbone, modality encoders, and common
training hyperparameters. The comparison therefore isolates the proposed
changes to recursive latent rollout and cross-modal alignment.

Because CoMM indicates that multimodal representation quality can depend on
augmentation strength \cite{dufumier2025comm}, we evaluate two random crop
settings. In \textsc{IPR-S}, the suffix S denotes stronger cropping, and the
crop retains between $10\%$ and $100\%$ of the original image area. In
\textsc{IPR-H}, the suffix H denotes higher image retention, and the crop
retains between $50\%$ and $100\%$, preserving more visual context. The
configurations differ only in the minimum retained image area and are used to
evaluate sensitivity to crop strength.

Training uses 500,000 environment steps for Run, Spin, and Walk and 300,000
steps for Catch and Swing-up. The first 1,000 agent decisions use random
actions. The replay buffer capacity is 50,000, and both the RL and auxiliary
batch sizes are 64. We use a discount factor of $\gamma=0.99$, a prediction horizon
$H=5$, and a learning rate of $10^{-3}$ for the actor,
critic, encoders and auxiliary modules. The prediction and alignment loss
weights are $\lambda_{\mathrm{pred}}=1$ and
$\lambda_{\mathrm{align}}=0.05$, respectively.
For RGB--Position, the action-prefix Transformer has two layers,
two attention heads, and embeddings of dimension 50. The alignment projector
outputs features of dimension 256 and uses an InfoNCE temperature of $0.1$.
The momentum update coefficient is $0.01$. Learned mask tokens are
initialised from a truncated normal distribution with standard deviation
$0.02$.
Hyperparameter values were selected empirically. Shared settings were then held fixed across the compared methods within
each observation setting. The policy is evaluated every 5,000 environment steps. Each evaluation
reports the mean return over five episodes.

\paragraph{Results.}
Tables~\ref{tab:dmc-depth-results} and
\ref{tab:dmc-position-results} report the mean final return and standard
deviation over five seeds. Figures~\ref{fig:dmc-depth-curves} and
\ref{fig:dmc-position-curves} show the corresponding learning curves.

\begin{table}[t]
    \centering
    \caption{Final returns under RGB--Depth. Values are mean final return
    $\pm$ standard deviation over five seeds. The highest mean in
    each row is shown in bold.}
    \small
    \setlength{\tabcolsep}{2pt}
    \begin{tabular}{@{}lccc@{}}
        \hline
        Task & \textsc{Base} & \textsc{IPR-S} & \textsc{IPR-H} \\
        \hline
        Catch
        & $817.3 \pm 92.6$
        & $\mathbf{936.5 \pm 9.0}$
        & $913.3 \pm 55.0$ \\
        Run
        & $490.8 \pm 65.6$
        & $510.0 \pm 82.9$
        & $\mathbf{572.3 \pm 26.3}$ \\
        Spin
        & $557.3 \pm 87.6$
        & $665.2 \pm 115.0$
        & $\mathbf{665.9 \pm 121.9}$ \\
        Swing-up
        & $735.8 \pm 76.5$
        & $728.9 \pm 69.4$
        & $\mathbf{767.2 \pm 46.5}$ \\
        Walk
        & $698.4 \pm 100.9$
        & $\mathbf{732.1 \pm 104.9}$
        & $710.8 \pm 94.5$ \\
        \hline
        \textit{Average}
        & $659.9$
        & $714.5$
        & $\mathbf{725.9}$ \\
        \hline
    \end{tabular}
    \label{tab:dmc-depth-results}
\end{table}
\begin{figure*}[t]
    \centering
    \setlength{\tabcolsep}{2pt}
    \begin{tabular}{@{}ccc@{}}
        \includegraphics[width=0.30\textwidth]
        {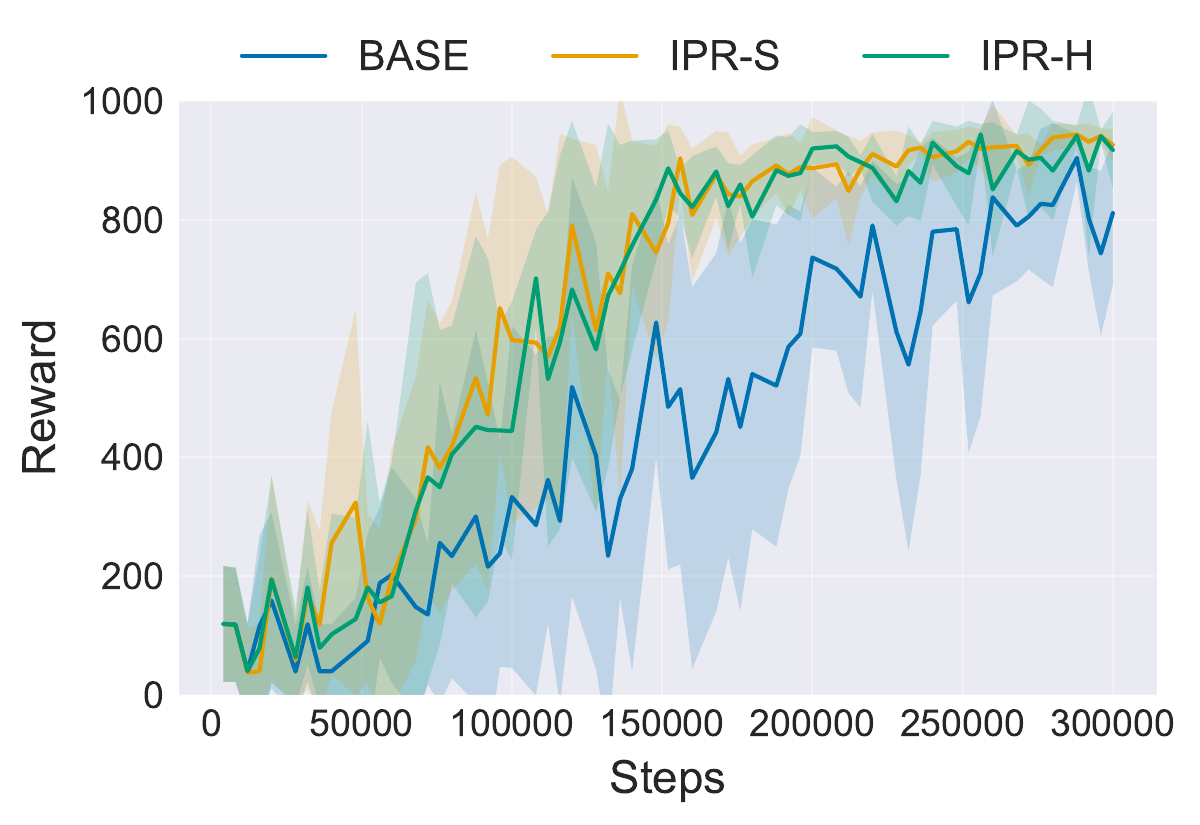} &
        \includegraphics[width=0.30\textwidth]
        {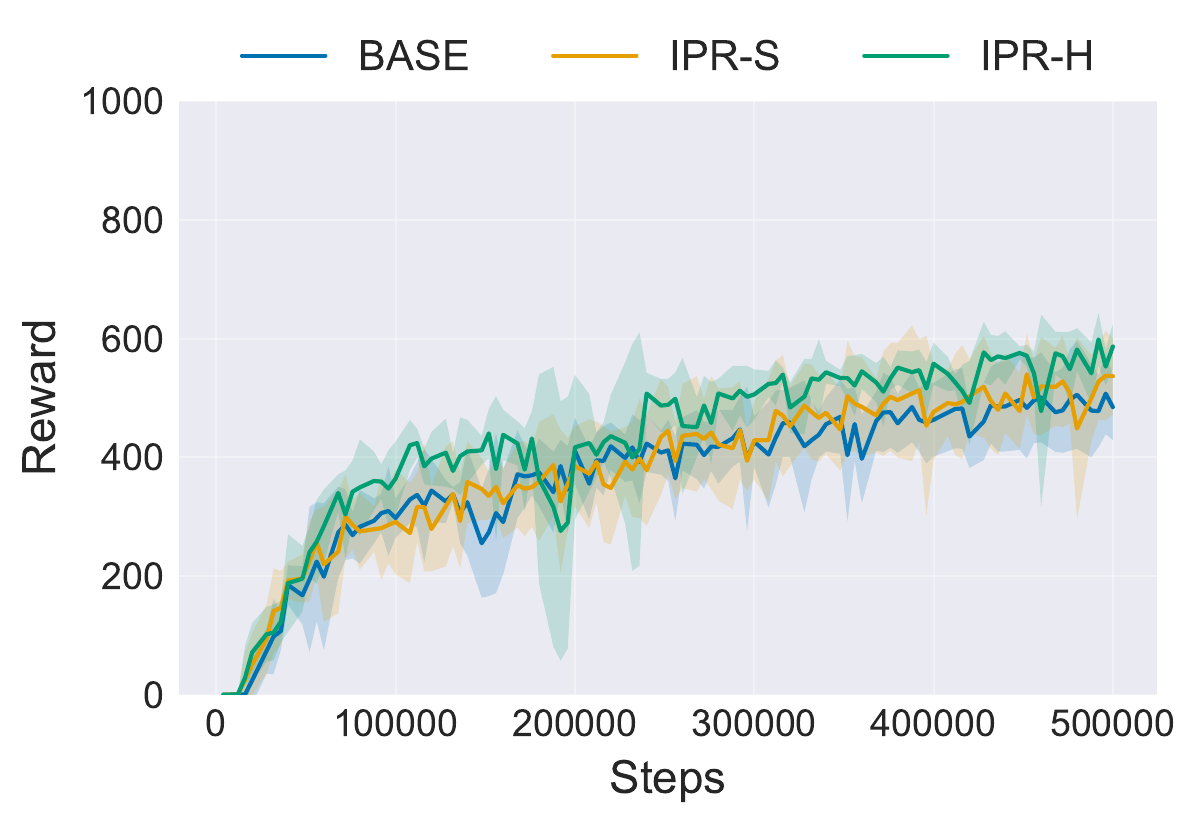} &
        \includegraphics[width=0.30\textwidth]
        {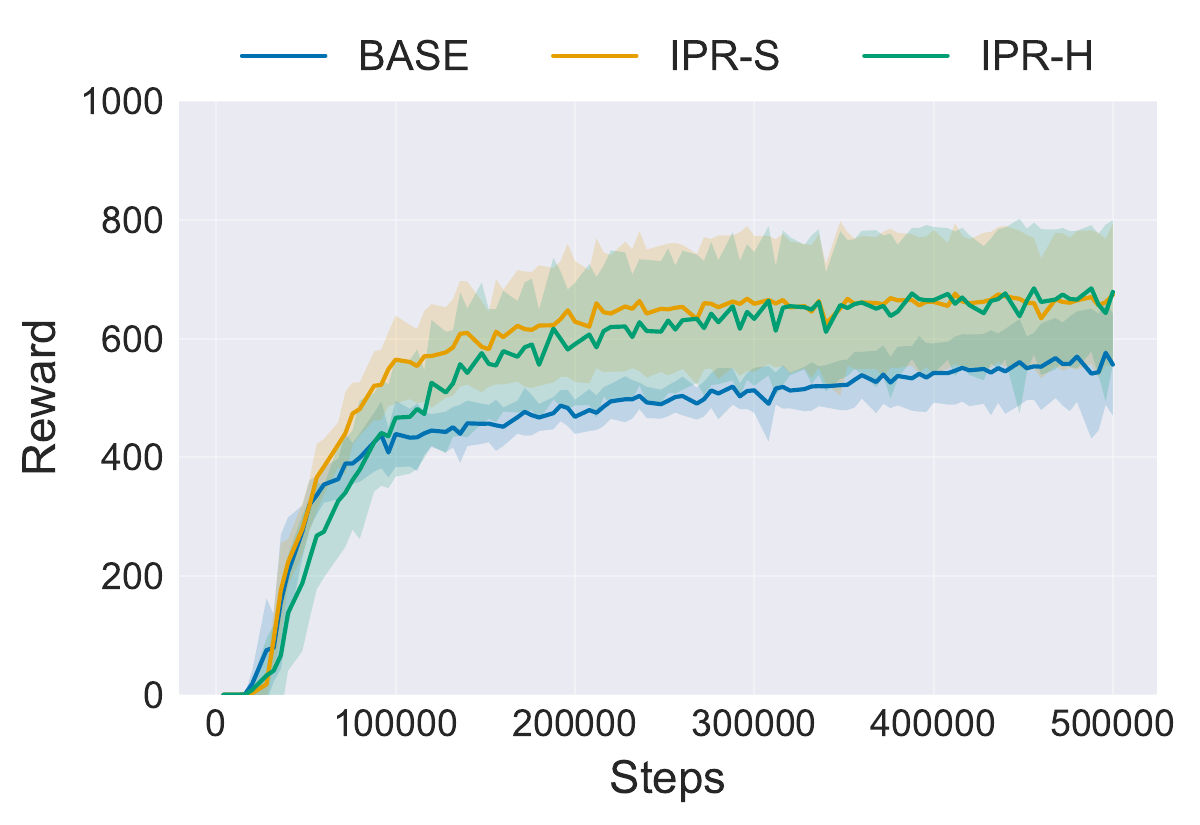} \\
        (a) Catch & (b) Run & (c) Spin \\[0.5mm]
        \multicolumn{3}{c}{
            \begin{tabular}{@{}cc@{}}
                \includegraphics[width=0.30\textwidth]
                {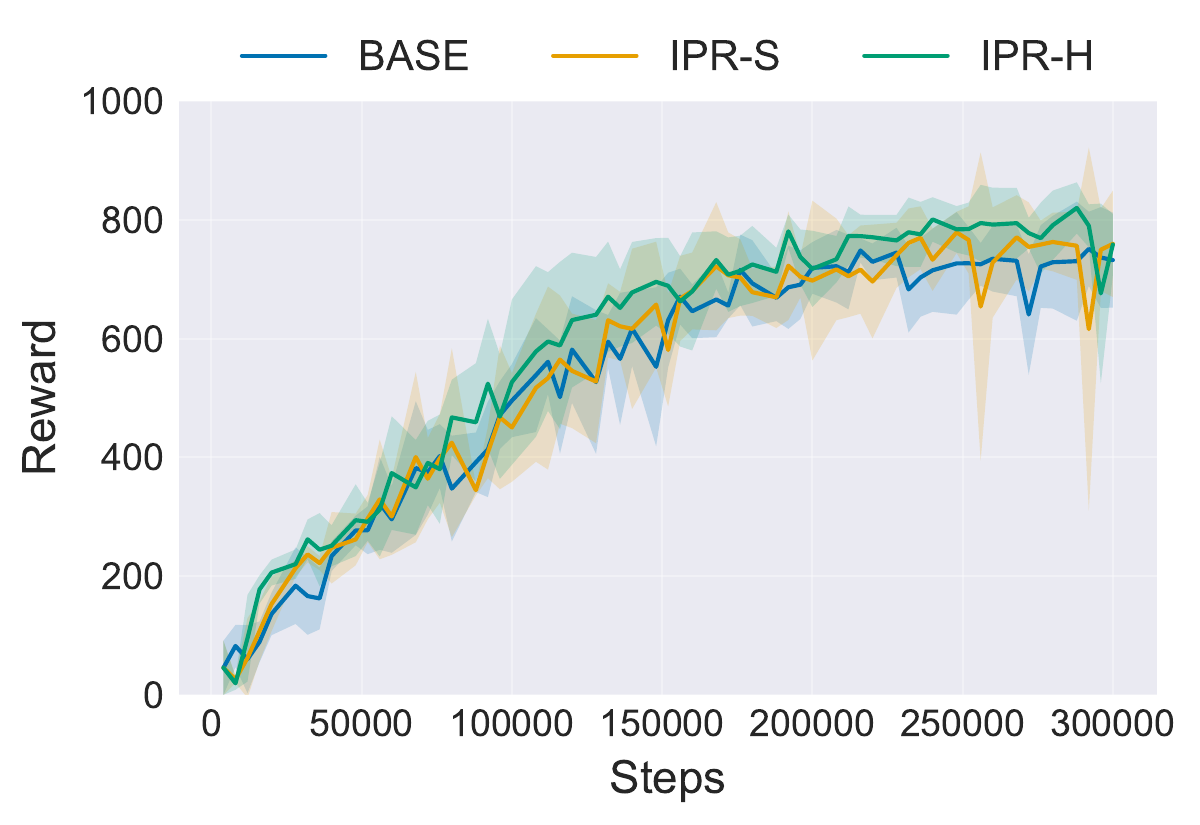} &
                \includegraphics[width=0.30\textwidth]
                {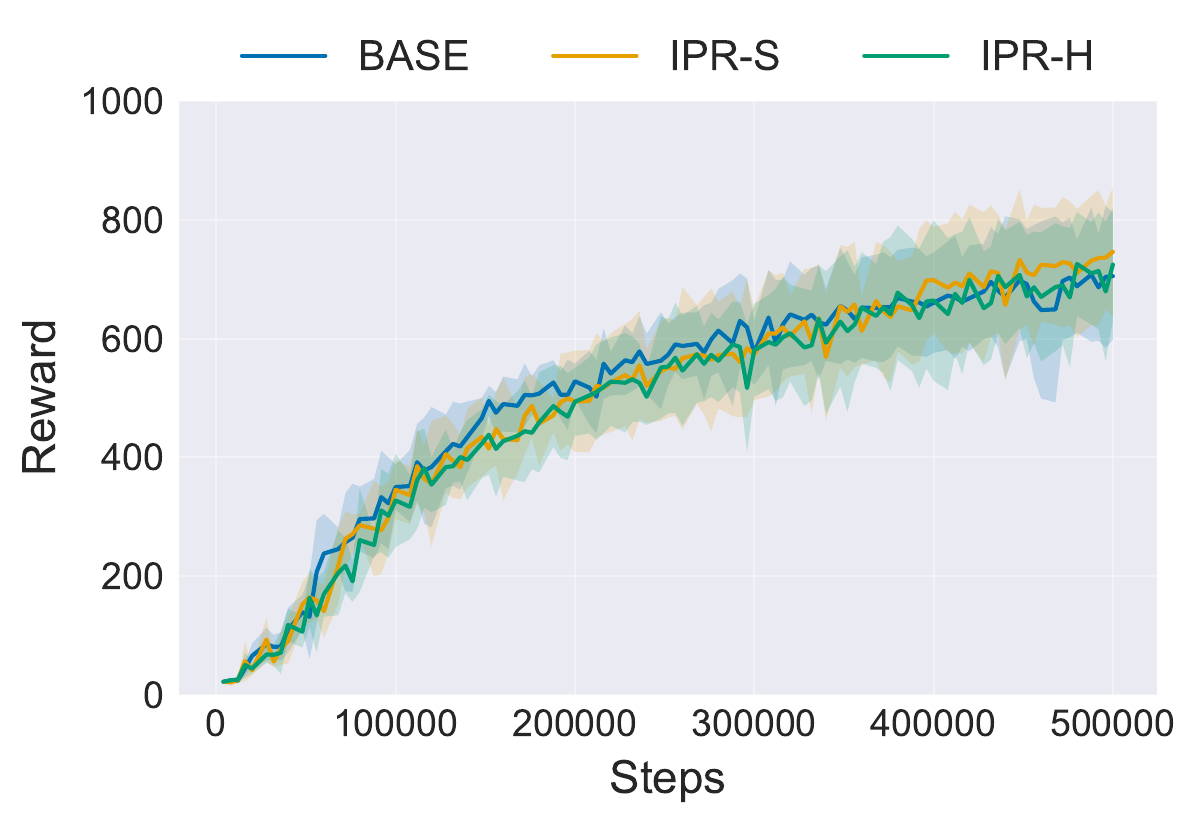} \\
                (d) Swing-up & (e) Walk
            \end{tabular}
        }
    \end{tabular}
    \caption{Learning curves on five DMC tasks under RGB--Depth.
    Solid lines show the mean over five seeds, and shaded regions show one
    standard deviation. Catch and Swing-up use 300,000 training steps, while
    the other tasks use 500,000.}
    \label{fig:dmc-depth-curves}
\end{figure*}
\begin{table*}[t]
    \centering
    \caption{Final returns and component ablations under RGB--Position.
    Values are mean final return $\pm$ standard deviation over five seeds.
    \textsc{Base-C} uses CoMM alignment with recursive latent rollout, while
    \textsc{Base-P} uses action-prefix prediction with the baseline alignment
    objective.
    The highest mean in each row is shown in bold.}
    \small
    \setlength{\tabcolsep}{4pt}
    \begin{tabular}{@{}lccccc@{}}
        \hline
        Task
        & \textsc{Base}
        & \textsc{Base-C}
        & \textsc{Base-P}
        & \textsc{IPR-S}
        & \textsc{IPR-H} \\
        \hline
        Catch
        & $940.2 \pm 20.3$
        & $950.7 \pm 9.8$
        & $943.4 \pm 12.7$
        & $958.8 \pm 3.8$
        & $\mathbf{960.6 \pm 5.7}$ \\
        Run
        & $563.2 \pm 39.2$
        & $588.7 \pm 23.5$
        & $559.6 \pm 33.2$
        & $596.2 \pm 97.2$
        & $\mathbf{597.2 \pm 64.6}$ \\
        Spin
        & $659.8 \pm 65.1$
        & $659.4 \pm 86.9$
        & $609.2 \pm 84.7$
        & $\mathbf{679.8 \pm 86.1}$
        & $660.1 \pm 90.3$ \\
        Swing-up
        & $771.2 \pm 52.8$
        & $774.8 \pm 47.6$
        & $782.4 \pm 67.7$
        & $785.8 \pm 56.3$
        & $\mathbf{822.7 \pm 23.0}$ \\
        Walk
        & $832.2 \pm 76.7$
        & $855.6 \pm 75.6$
        & $825.0 \pm 64.1$
        & $\mathbf{868.1 \pm 33.1}$
        & $855.5 \pm 75.0$ \\
        \hline
        \textit{Average}
        & $753.3$
        & $765.8$
        & $743.9$
        & $777.7$
        & $\mathbf{779.2}$ \\
        \hline
    \end{tabular}
    \label{tab:dmc-position-results}
\end{table*}

\begin{figure*}[t]
    \centering
    \setlength{\tabcolsep}{2pt}
    \begin{tabular}{@{}ccc@{}}
        \includegraphics[width=0.30\textwidth]
        {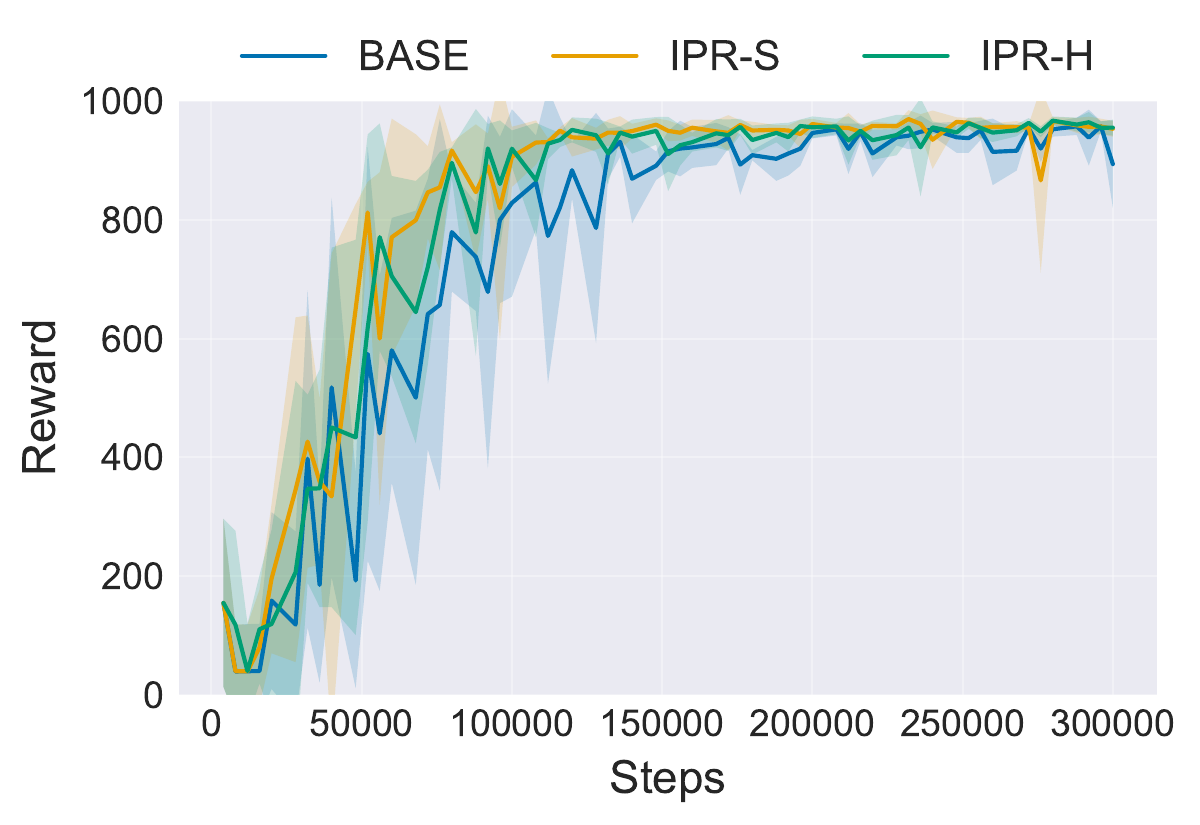} &
        \includegraphics[width=0.30\textwidth]
        {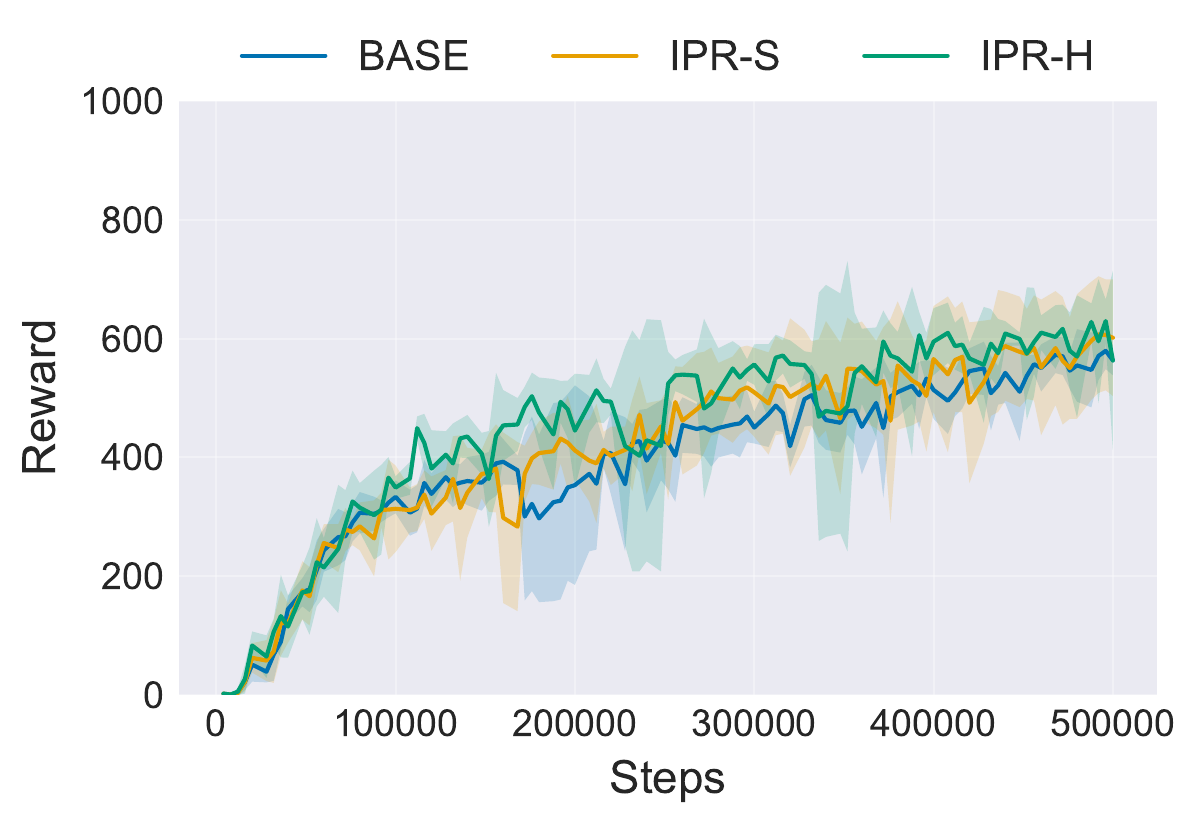} &
        \includegraphics[width=0.30\textwidth]
        {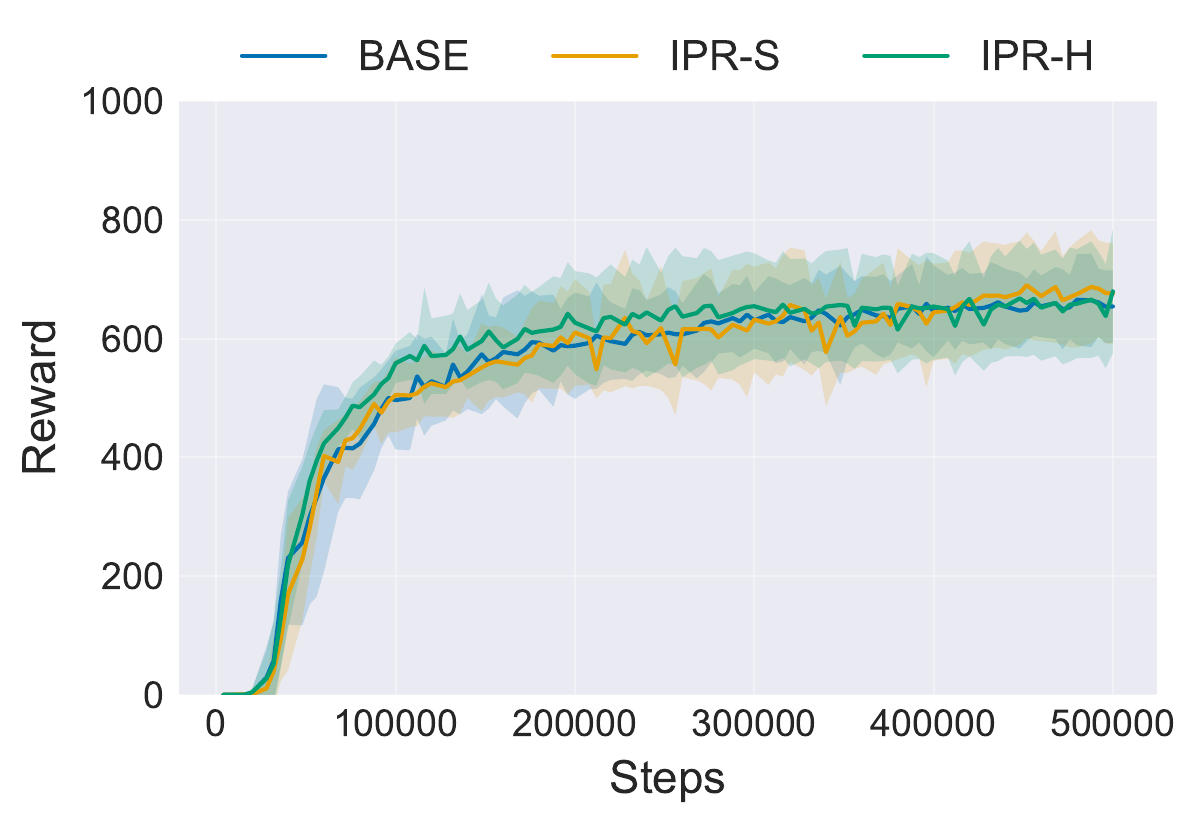} \\
        (a) Catch & (b) Run & (c) Spin \\[0.5mm]
        \multicolumn{3}{c}{
            \begin{tabular}{@{}cc@{}}
                \includegraphics[width=0.30\textwidth]
                {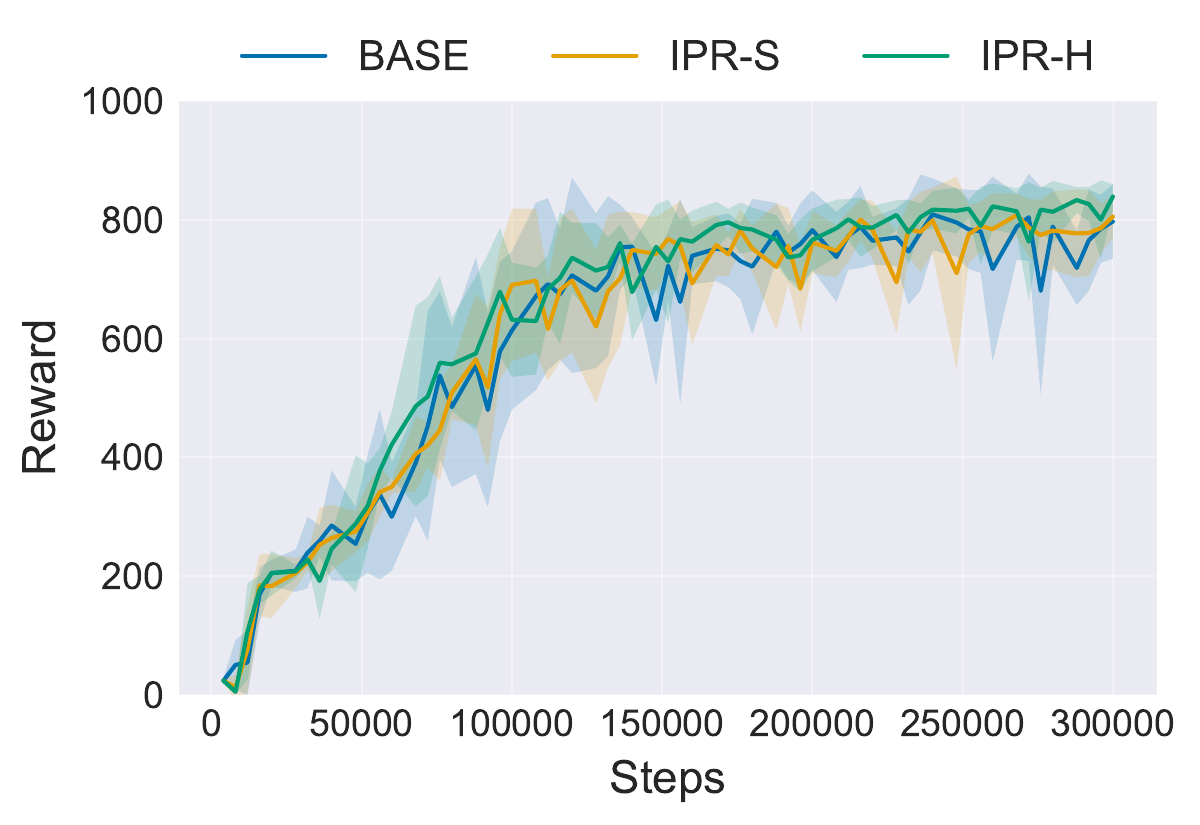} &
                \includegraphics[width=0.30\textwidth]
                {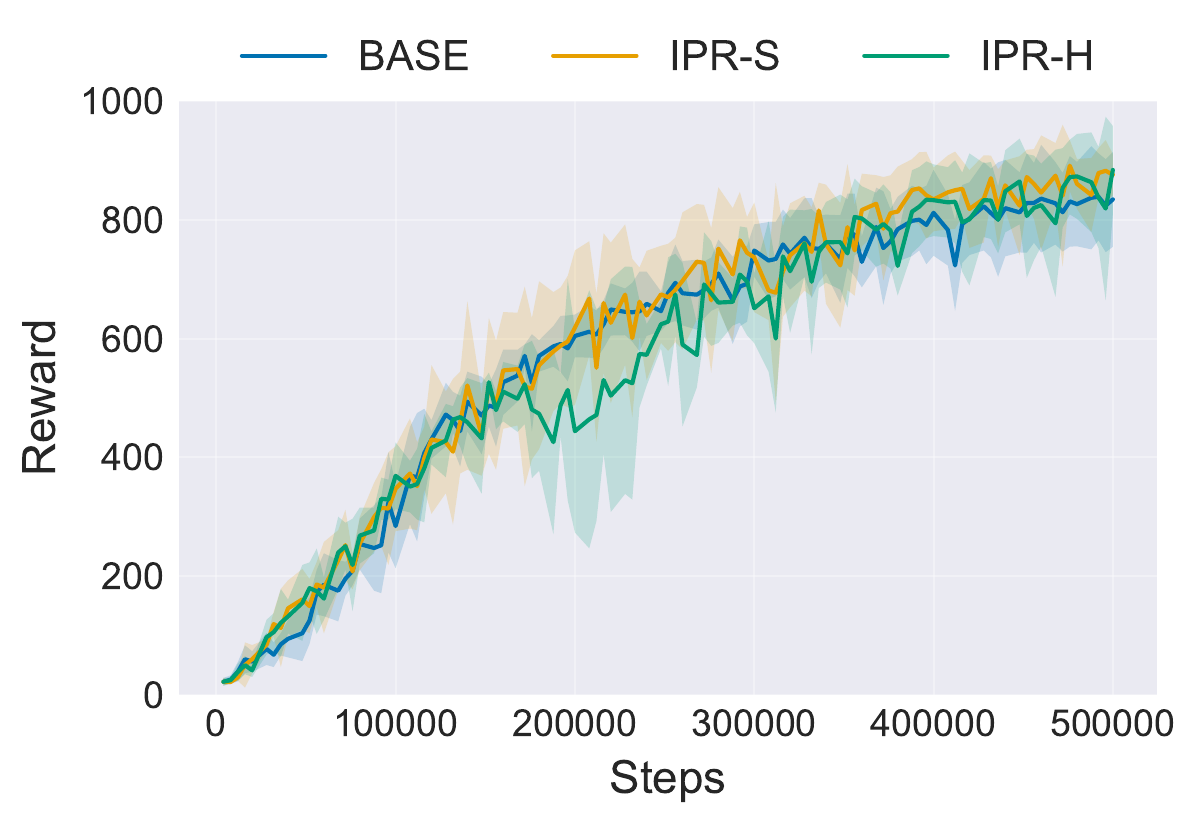} \\
                (d) Swing-up & (e) Walk
            \end{tabular}
        }
    \end{tabular}
    \caption{Learning curves on five DMC tasks under RGB--Position. Solid
    lines show the mean over five seeds, and shaded regions show one standard
    deviation. Catch and Swing-up use 300,000 training steps, while the other
    tasks use 500,000.}
    \label{fig:dmc-position-curves}
\end{figure*}

Under RGB--Depth, both IPR configurations achieve higher averages across
the five tasks than \textsc{Base}. \textsc{IPR-H} obtains the highest average,
increasing it from $659.9$ to $725.9$, while \textsc{IPR-S} reaches $714.5$.

Under RGB--Position, both IPR configurations also improve the average,
although the gains are smaller. \textsc{IPR-H} reaches $779.2$, followed
closely by \textsc{IPR-S} at $777.7$, compared with $753.3$ for
\textsc{Base}.

\subsection{Ablation Study}
\label{sec:abla}
We isolate the two changes made to MJPR in IPR using RGB--Position
observations. The \textsc{Base}, \textsc{Base-C}, \textsc{Base-P}, and
\textsc{IPR-S} columns in Table~\ref{tab:dmc-position-results} form a
factorial comparison between recursive and action-prefix prediction and
between the baseline and CoMM alignment configurations. \textsc{Base-C} and
\textsc{IPR-S} use the same random resized crop setting, with the retained
image area sampled between $10\%$ and $100\%$.
\textsc{IPR-H} is reported separately to compare the two crop settings within the complete IPR method. All methods use the same optimisation procedure, prediction horizon, evaluation protocol, and five
seeds.

\paragraph{Results.}
Replacing the baseline alignment objective with anchor-level CoMM alignment while retaining the recursive latent rollout increases the mean final return on Catch,
Run, Swing-up, and Walk, while producing a similar result on Spin. The
average across the five tasks for \textsc{Base-C} is $1.7\%$ higher than that of
\textsc{Base}.

The effect of action-prefix prediction differs between the two alignment
configurations. With the baseline alignment objective, replacing the
recursive latent rollout with action-prefix prediction reduces the
average across the five tasks from $753.3$ for \textsc{Base} to $743.9$
for \textsc{Base-P}. With CoMM alignment and the same minimum retained
image area ratio of $0.1$, action-prefix prediction increases the average
from $765.8$ for \textsc{Base-C} to $777.7$ for \textsc{IPR-S}. This observed reversal is consistent with an interaction
between the prediction and alignment configurations, although the
variation across seeds prevents a stronger statistical conclusion.

We also compare the end-to-end training wall time of \textsc{Base} and
\textsc{Base-P} under RGB--Depth. This comparison examines the computational
effect of replacing the recursive latent rollout with action-prefix prediction while
retaining the baseline alignment objective. Both methods use the same
training budget, and each job uses one NVIDIA A100 graphics processing unit,
four central processing unit cores, and 32 gigabytes of memory. Across the
five tasks, \textsc{Base-P} reduces the mean wall time by $11.4\%$ to
$16.9\%$. The average reduction is $13.0\%$, corresponding to a speedup of
approximately $1.15\times$.

The timing analysis includes all successful job records. Wall time includes
environment interaction, evaluation, and data processing, so it measures
total training time rather than the prediction module alone.

\subsection{Webots Environment}
\label{sec:webot}

\paragraph{Environment Setting.}

We construct a navigation environment in Webots R2025a
\cite{michel2004webots} using an E-puck
differential-drive robot. The arena contains four static cubic obstacles and
three red spherical targets. At the beginning of each episode, the robot,
obstacles, and targets are randomly positioned subject to minimum distance
constraints. An episode terminates when the robot collides with an obstacle
or boundary, or after 200 interaction steps. Reaching a target does not end
the episode. Instead, the collected target is relocated to another valid
position.

\begin{figure}[t]
    \centering
    \includegraphics[width=0.72\columnwidth]{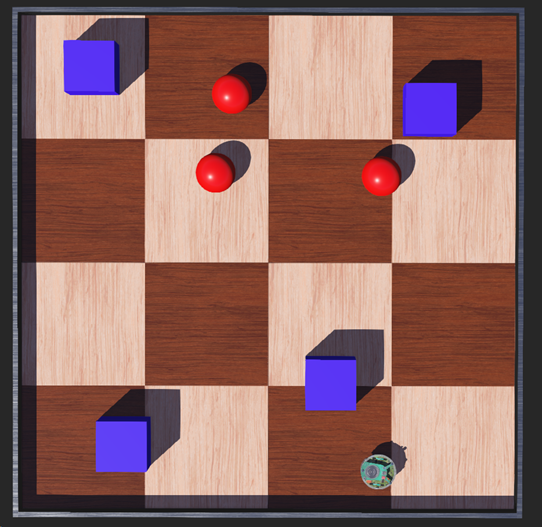}
    \caption{Static Webots navigation arena used in both observation
    settings.}
    \label{fig:webots-env}
\end{figure}

We evaluate two observation settings. The setting with three modalities
contains an $84\times84$ single-channel camera image, 360 LiDAR measurements,
and a five-dimensional position vector
$(x,y,\theta,\Delta x,\Delta y)$ containing the planar position, orientation,
and displacement of the robot. Both settings use the same Webots world,
object placement, environment dynamics, reset procedure, actions, rewards,
and termination conditions. The setting with two modalities differs only by
removing LiDAR and retaining the camera image and position vector.

The camera image is obtained by averaging the colour channels and scaling
the result to $[0,1]$. Invalid LiDAR measurements in the setting with three
modalities are replaced with zero. The displacement components represent the change in robot position
between consecutive observations and are scaled by a factor of 50.
The continuous action contains separate commands for the left and right
wheels, with each command ranging from $-1$ to $1$. The commands are linearly
scaled to nonnegative wheel angular velocities, where $-1$ corresponds to a
stationary wheel and $1$ corresponds to the maximum wheel velocity. Each
action is executed for 20 basic Webots simulation steps.

The agent receives the rewards listed in
Table~\ref{tab:webots-reward}. The wheel-control term is applied at every
interaction step, while target and collision rewards are added when the
corresponding events occur.

\begin{table}[t]
    \centering
    \caption{Reward components used in the Webots environment.
    The wheel-control term acts as a turning penalty by discouraging
    asymmetric wheel commands.}
    \small
    \setlength{\tabcolsep}{4pt}
    \begin{tabular}{@{}p{0.70\columnwidth}r@{}}
        \hline
        Event or condition & Reward \\
        \hline
        Reaching a target & $+10$ \\
        Collision with an obstacle or boundary & $-10$ \\
        Wheel control at each step
        & $0.1a_{t,L}a_{t,R}-0.1$ \\
        \hline
    \end{tabular}
    \label{tab:webots-reward}
\end{table}

We compare \textsc{Base}, \textsc{IPR-S}, and \textsc{IPR-H} under both
observation settings. The two IPR configurations use minimum retained image area
ratios of $0.1$ and $0.5$, respectively. All methods are trained for 51,000
interaction steps, with random actions used during the first 1,000 steps.
The prediction and alignment loss
weights are $\lambda_{\mathrm{pred}}=1$ and
$\lambda_{\mathrm{align}}=0.1$, respectively. Evaluation is conducted every 1,000 steps using five episodes
and a deterministic policy.
\paragraph{Results.}

\begin{figure*}[t]
    \centering
    \setlength{\tabcolsep}{3pt}
    \begin{tabular}{@{}cc@{}}
        \includegraphics[width=0.47\textwidth]
        {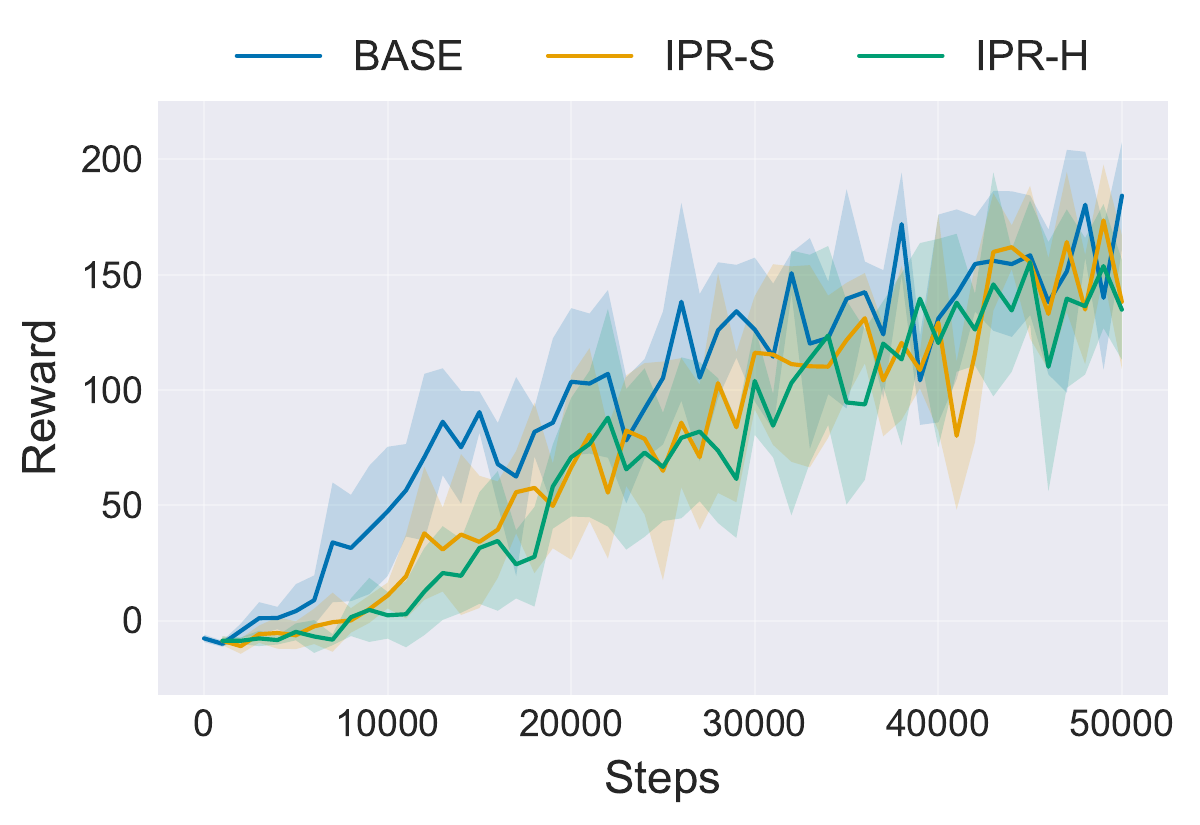}
        &
        \includegraphics[width=0.47\textwidth]
        {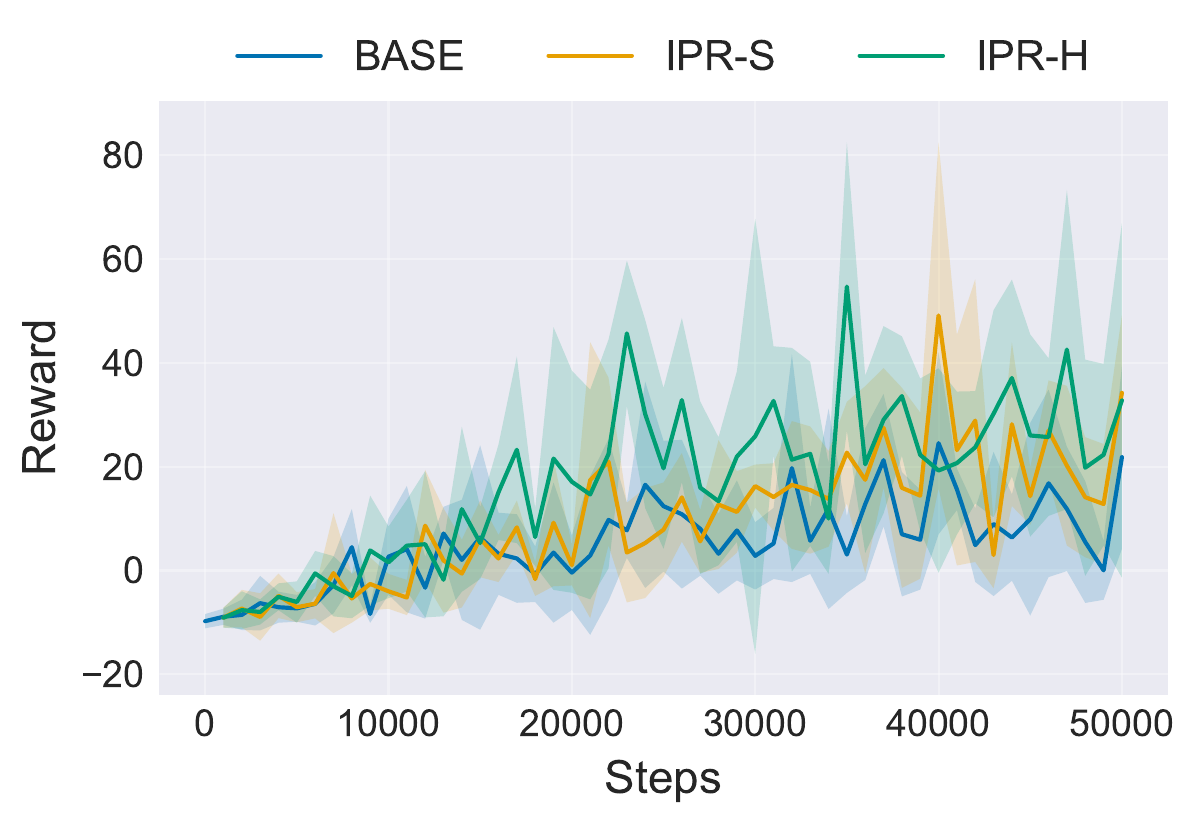}
        \\
        (a) Camera, LiDAR, and position
        &
        (b) Camera and position
    \end{tabular}
    \caption{Learning curves in the static Webots environment using
    (a) camera, LiDAR, and position observations and (b) camera and position
    observations. Solid lines show the mean over five seeds, and shaded
    regions show one standard deviation. The plots use different vertical
    scales because removing LiDAR substantially reduces the return.}
    \label{fig:webots-curves}
\end{figure*}

Figure~\ref{fig:webots-curves} presents the learning curves, while
Table~\ref{tab:webots-results} reports the mean and standard deviation over
the five seeds
$\{1000, 2000, 3000, 4000, 5000\}$.

\begin{table}[t]
    \centering
    \caption{Final Webots returns with three and two input modalities.
    Values are mean final return $\pm$ standard deviation over five seeds.
    The highest mean in each row is shown in bold.}
    \label{tab:webots-results}
    \small
    \setlength{\tabcolsep}{2pt}
    \begin{tabular}{@{}lccc@{}}
        \hline
        Modalities & \textsc{Base} & \textsc{IPR-S} & \textsc{IPR-H} \\
        \hline
        \shortstack[l]{Camera, LiDAR,\\ position}
        & $\mathbf{158.9{\pm}18.3}$
        & $148.7{\pm}6.6$
        & $134.9{\pm}17.0$ \\
        \shortstack[l]{Camera and\\ position}
        & $11.4{\pm}11.9$
        & $21.7{\pm}6.5$
        & $\mathbf{28.6{\pm}8.8}$ \\
        \hline
    \end{tabular}
\end{table}

The method ranking changes between the two observation settings. With three
modalities,
\textsc{Base} improves earlier and obtains the highest mean final return. The
mean final returns of \textsc{IPR-S} and \textsc{IPR-H} are $6.4\%$ and $15.1\%$ below
\textsc{Base}, respectively.

Removing LiDAR reduces the returns of all methods but reverses their ranking.
With two modalities, both IPR configurations obtain higher mean final returns than
\textsc{Base},
with \textsc{IPR-H} producing the highest mean final return.

% ------------------------------------------------------------------
% 5. Discussion
% ------------------------------------------------------------------
\section{Discussion}
\label{sec:discussion}

The DMC results show that the effect of IPR depends on both the observation
setting and the crop configuration. Although \textsc{IPR-H} achieves the
highest average under both observation settings, \textsc{IPR-S}
performs better on some individual tasks. Preserving more image content is
therefore beneficial on average but is not uniformly preferable. This
variation suggests that crop strength affects the balance between retaining
task-relevant information and introducing useful variation during
representation learning.

The ablation results provide further evidence that the two IPR modifications
should not be considered independently. Replacing CURL with
CoMM alignment gives the more consistent improvement, whereas
action-prefix prediction does not consistently improve return when CURL is
retained. Its effect is more favourable when combined with CoMM alignment in
the complete IPR framework. This pattern suggests that the benefit of
action-prefix prediction may depend on how representations are aligned.
However, the current results describe this interaction only through observed
mean returns and do not establish its statistical reliability.

The Webots results also depend on the information provided by the observation
modalities. Camera, LiDAR, and position observations may already supply
sufficient and partly redundant information for navigation. In particular,
LiDAR directly measures obstacle distance, so the additional prediction and
alignment objectives may provide limited information beyond that available
to \textsc{Base}.

Without LiDAR, the agent must infer distances to obstacles from the camera
and position observations. Both IPR configurations achieve higher mean final
returns than \textsc{Base} in this setting, suggesting that predictive and cross-modal supervision may be
more useful when direct range information is unavailable. This explanation
is consistent with the observed results but remains a hypothesis. Sensor and
loss ablations are required to determine whether missing range information
causes the change in method ranking.

Fast LeWorldModel shows that action-prefix prediction permits prediction
horizons to be computed in parallel \cite{gao2026fastlewm}. The lower
end-to-end training wall time of \textsc{Base-P} is consistent with this
motivation, although the experiment does not isolate the prediction module.

Finally, several tasks exhibit substantial variation across seeds, and no
formal hypothesis test is performed. The reported comparisons therefore
support conclusions about observed trends in mean return rather than
statistical significance or general superiority. Additional runs, formal
statistical tests, and experiments across further environments are needed to
determine the reliability and scope of these findings.

% ------------------------------------------------------------------
% 6. Conclusion
% ------------------------------------------------------------------
\section{Conclusion}
\label{sec:conclusion}

This work integrates action-prefix prediction with anchor-level cross-modal
alignment for online multimodal reinforcement learning. IPR achieves higher averages than \textsc{Base} in the evaluated DMC settings, while
the Webots results indicate that its effect depends on the information
provided by the observation modalities. These findings do not establish
general superiority or a causal explanation for the observed differences.

Future work should test the proposed explanation through targeted sensor and
loss ablations, use additional runs with formal statistical tests, evaluate
further robotic tasks, and examine transfer to real systems.

\section*{Acknowledgments}

The authors acknowledge the use of the University of New South Wales (UNSW)
Compute Cluster Katana for the experiments reported in this work under Digital Object Identifier
10.26190/669X-A286.

\balance
\bibliographystyle{named}
\bibliography{references}

@inproceedings{haarnoja2018sac,
  title     = {Soft Actor-Critic: Off-Policy Maximum Entropy Deep
               Reinforcement Learning with a Stochastic Actor},
  author    = {Haarnoja, Tuomas and Zhou, Aurick and Abbeel, Pieter and
               Levine, Sergey},
  booktitle = {Proceedings of the 35th International Conference on
               Machine Learning},
  series    = {Proceedings of Machine Learning Research},
  volume    = {80},
  pages     = {1861--1870},
  publisher = {PMLR},
  year      = {2018},
  url       = {https://proceedings.mlr.press/v80/haarnoja18b.html}
}

@inproceedings{laskin2020rad,
  title     = {Reinforcement Learning with Augmented Data},
  author    = {Laskin, Misha and Lee, Kimin and Stooke, Adam and
               Pinto, Lerrel and Abbeel, Pieter and Srinivas, Aravind},
  booktitle = {Advances in Neural Information Processing Systems},
  volume    = {33},
  year      = {2020},
  url       = {https://proceedings.neurips.cc/paper/2020/hash/e615c82aba461681ade82da2da38004a-Abstract.html}
}

@inproceedings{yarats2021drq,
  title     = {Image Augmentation Is All You Need: Regularizing Deep
               Reinforcement Learning from Pixels},
  author    = {Kostrikov, Ilya and Yarats, Denis and Fergus, Rob},
  booktitle = {International Conference on Learning Representations},
  year      = {2021},
  url       = {https://openreview.net/forum?id=GY6-6sTvGaf}
}

@inproceedings{laskin2020curl,
  title     = {{CURL}: Contrastive Unsupervised Representations for
               Reinforcement Learning},
  author    = {Laskin, Michael and Srinivas, Aravind and Abbeel, Pieter},
  booktitle = {Proceedings of the 37th International Conference on
               Machine Learning},
  series    = {Proceedings of Machine Learning Research},
  volume    = {119},
  pages     = {5639--5650},
  publisher = {PMLR},
  year      = {2020},
  url       = {https://proceedings.mlr.press/v119/laskin20a.html}
}

@article{oord2018cpc,
  title   = {Representation Learning with Contrastive Predictive Coding},
  author  = {van den Oord, Aaron and Li, Yazhe and Vinyals, Oriol},
  journal = {arXiv preprint arXiv:1807.03748},
  year    = {2018},
  url     = {https://arxiv.org/abs/1807.03748}
}

@inproceedings{stooke2021atc,
  title     = {Decoupling Representation Learning from Reinforcement Learning},
  author    = {Stooke, Adam and Lee, Kimin and Abbeel, Pieter and
               Laskin, Michael},
  booktitle = {Proceedings of the 38th International Conference on
               Machine Learning},
  series    = {Proceedings of Machine Learning Research},
  volume    = {139},
  pages     = {9870--9879},
  publisher = {PMLR},
  year      = {2021},
  url       = {https://proceedings.mlr.press/v139/stooke21a.html}
}

@inproceedings{schwarzer2021spr,
  title     = {Data-Efficient Reinforcement Learning with
               Self-Predictive Representations},
  author    = {Schwarzer, Max and Anand, Ankesh and Goel, Rishab and
               Hjelm, R. Devon and Courville, Aaron and Bachman, Philip},
  booktitle = {International Conference on Learning Representations},
  year      = {2021},
  url       = {https://openreview.net/forum?id=uCQfPZwRaUu}
}

@inproceedings{chen2020simclr,
  title     = {A Simple Framework for Contrastive Learning of Visual
               Representations},
  author    = {Chen, Ting and Kornblith, Simon and Norouzi, Mohammad and
               Hinton, Geoffrey},
  booktitle = {Proceedings of the 37th International Conference on
               Machine Learning},
  series    = {Proceedings of Machine Learning Research},
  volume    = {119},
  pages     = {1597--1607},
  publisher = {PMLR},
  year      = {2020},
  url       = {https://proceedings.mlr.press/v119/chen20j.html}
}

@inproceedings{tian2020cmc,
  title     = {Contrastive Multiview Coding},
  author    = {Tian, Yonglong and Krishnan, Dilip and Isola, Phillip},
  booktitle = {Computer Vision--ECCV 2020},
  pages     = {776--794},
  publisher = {Springer},
  year      = {2020},
  doi       = {10.1007/978-3-030-58621-8_45}
}

@inproceedings{vaswani2017attention,
  title     = {Attention Is All You Need},
  author    = {Vaswani, Ashish and Shazeer, Noam and Parmar, Niki and
               Uszkoreit, Jakob and Jones, Llion and Gomez, Aidan N. and
               Kaiser, Lukasz and Polosukhin, Illia},
  booktitle = {Advances in Neural Information Processing Systems},
  volume    = {30},
  pages     = {5998--6008},
  year      = {2017}
}

@inproceedings{wang2025mjpr,
  title     = {{MJPR}: Multi-Modal Joint Predictive Representation in
               Deep Reinforcement Learning},
  author    = {Wang, Zehan and He, Ziming and Wang, Zijia and He, Hua and
               Yang, Beiya and Shi, Haobin},
  booktitle = {2025 IEEE International Conference on Robotics and
               Automation (ICRA)},
  pages     = {1--7},
  year      = {2025},
  doi       = {10.1109/ICRA55743.2025.11128137}
}

@inproceedings{dufumier2025comm,
  title     = {What to Align in Multimodal Contrastive Learning?},
  author    = {Dufumier, Benoit and Castillo-Navarro, Javiera and
               Tuia, Devis and Thiran, Jean-Philippe},
  booktitle = {International Conference on Learning Representations},
  year      = {2025},
  url       = {https://openreview.net/forum?id=Pe3AxLq6Wf}
}

@article{gao2026fastlewm,
  title   = {Fast {LeWorldModel}},
  author  = {Gao, Yuntian and Xu, Xiangyu},
  journal = {arXiv preprint arXiv:2606.26217},
  year    = {2026},
  doi     = {10.48550/arXiv.2606.26217},
  url     = {https://arxiv.org/abs/2606.26217}
}

@book{sutton2018reinforcement,
  title     = {Reinforcement Learning: An Introduction},
  author    = {Sutton, Richard S. and Barto, Andrew G.},
  edition   = {2},
  publisher = {MIT Press},
  address   = {Cambridge, Massachusetts},
  year      = {2018},
  url       = {https://mitpress.mit.edu/9780262039246/reinforcement-learning/}
}

@article{tassa2018deepmind,
  title   = {{DeepMind Control Suite}},
  author  = {Tassa, Yuval and Doron, Yotam and Muldal, Alistair and
             Erez, Tom and Li, Yazhe and de Las Casas, Diego and
             Budden, David and Abdolmaleki, Abbas and Merel, Josh and
             Lefrancq, Andrew and Lillicrap, Timothy and Riedmiller, Martin},
  journal = {arXiv preprint arXiv:1801.00690},
  year    = {2018},
  doi     = {10.48550/arXiv.1801.00690},
  url     = {https://arxiv.org/abs/1801.00690}
}

@article{michel2004webots,
  title   = {{Cyberbotics Ltd. Webots}: Professional Mobile Robot Simulation},
  author  = {Michel, Olivier},
  journal = {International Journal of Advanced Robotic Systems},
  volume  = {1},
  number  = {1},
  pages   = {39--42},
  year    = {2004},
  doi     = {10.5772/5618}
}

\end{document}